\documentclass[11pt,a4paper]{article}
\usepackage[hyperref]{AACL-IJCNLP2020}
\usepackage{times}
\usepackage{latexsym}

\usepackage{microtype}

\aclfinalcopy % Uncomment this line for the final submission
\usepackage{mathtools}
\usepackage{algorithm,algorithmic}
\usepackage{framed}
\usepackage{mathrsfs}  
\usepackage{graphicx}
\usepackage{amssymb}
\usepackage{booktabs}
\newcommand{\purple}{\color{purple}}

\newcommand{\diag}{\text{diag}}
\rmfamily
\sffamily
\ttfamily
\usepackage{caption}
\usepackage{subfigure}
\usepackage{graphicx}
\usepackage{CJKutf8}
\usepackage{titlesec}
\usepackage{setspace}

\usepackage{verbatim}
\usepackage{latexsym} 
\usepackage{xcolor}

\usepackage{soul}
\usepackage{url}

\title{\textsf{FILM}: A \textsf{F}ast, \textsf{I}nterpretable, and \textsf{L}ow-rank \textsf{M}etric Learning Approach for Sentence Matching}

\author{Xiangru Tang \\
  Institute of Computing Technology \\
  Chinese Academy of Sciences  \\
  \texttt{xrtang@mails.ucas.ac.cn} \\\And
  Alan Aw\thanks{These authors contributed equally.} \\
  Department of Mathematics \\
  UC Berkeley \\
  \texttt{alanaw1@berkeley.edu} \\}

\begin{document}
\maketitle
\begin{abstract}
Detection of semantic similarity plays a vital role in sentence matching. It requires to learn discriminative representations of natural language. Recently, owing to more and more sophisticated model architecture, impressive progress has been made, along with a time-consuming training process and not-interpretable inference.
% In sentence matching and semantic analysis, detecting semantic similarity is a challenge that requires learning discriminative representations of natural language. Recent advances in the deep neural network enable us to learn semantic representation, but are getting time-consuming and fail in interpretation.
To alleviate this problem, we explore a metric learning approach, named \textsf{FILM} (\textsf{F}ast, \textsf{I}nterpretable, and \textsf{L}ow-rank \textsf{M}etric learning) to efficiently find a high discriminative projection of the high-dimensional data. We construct this metric learning problem as a manifold optimization problem, and solve it with the Cayley transformation method with Barzilai-Borwein step size.
% To alleviate this problem, in this paper we construct sentence matching as a manifold optimization problem that learns a distance function between sentences.
% % and obtain the semantic representation by learning a similarity or distance function.
% We explore a metric learning approach, named \textsf{FILM} (\textsf{F}ast, \textsf{I}nterpretable and \textsf{L}ow-rank \textsf{M}etric learning) to efficiently find a high discriminative projection of the high-dimensional data.
% that still preserves high discriminative power.
% To this end, our manifold optimization method is solved by the Cayley transformation method with Barzilai-Borwein step size. 
In experiments, we apply \textsf{FILM} with triplet loss minimization objective to the Quora Challenge and Semantic Textual Similarity (STS) Task. The results demonstrate that the \textsf{FILM} method achieves a superior performance as well as the fastest computation speed, which is consistent with our theoretical analysis of time complexity.
\end{abstract}

\section{Introduction}
\begin{comment}
\textsf{FILM}

shortcoming: 1.the whole paper does not show the advantage(Fast、interpretable) clearly
没特别说清楚 现在特别的优势

1.1 强调快Fast：需要理论证明复杂度实验对比说明快，并需要和深度学习方法对比（adam）

1.2 白盒：需要加特别的说明文字，模型是线性的，所以白盒

2.实验不够 experiment is not enough for text matching

2.1 add dataset：Yahoo! 

2.2加对比：other deep models, one experiment aims at proving our model is fastest, the other experiment aims at showing metric leanring能展现出来x初始的feature映射到的是什么地方，

和找表示的方法对比：text matching加数据集、
(什么情况下优势会明显，代表性不一样的data)

很多end2end的方法找表示，现在是KNN做任务

每一个X，Y的表示找到了

应该是有个思想：快、可解释，

解决方法：我们冲着matching任务来说，相似性是什么，是X的低维表示，可以是非常简单，metric learning，
以往有做的，算得快..（要跟）

在后面除了实验部分完善：1、两个数据集、两个数据集实验结果是完整的

2、强调快的实验

3、可解释的实验：metric leanring能展现出来映射到的是什么地方，x初始的feature

线性的embed出来的Y谁和谁更相关是可以展现的，在一些例子上是可以理解的

Y线性组合学出来的是哪些词组成的QA的表示，为什么match在一起是因为哪些词，（展现出L）比如高亮的方式是可以理解的，而不是用deep learn已经看不出来

简单可解释性TFIDF

缺点：
1.motivation不明确，contribution在哪
2.不完整

前面明确、后面完整、算法不是提的

\end{comment}

%text matching有很多deep learning方法，效果不错，但是可解释差，且速度不高(text 

Sentence Matching aims at comparing two sentences and identifying the semantic relationship, which serves as the core of many tasks such as information retrieval and semantic analysis. As a fundamental technology, sentence matching has broad applications, such as information retrieval, question answering, and dialog system. Among them, the core challenge is the semantic similarity of a sentence pair~\citep{wan2016match}, which could be calculated once we have the word representations. Recently, sentence matching has achieved significant progress with the development of neural networks~\cite{pang2016text,wang2017bilateral}.

%text matching就是要找相似度，只不过深度学习focus单独怎么表示，两个在一起怎么表示，所以速度很慢，且不知道找到了什么)

The neural networks represent two sentences individually to a dense vector in the same embedding space, and then define different functions to calculate the matching degree of the two-sentence vectors. However, they are getting extremely time-consuming as the networks are becoming more sophisticated and introducing more parameters. Even worse, it is still a black box for researchers and practitioners, and in urgent need of interpretability. We can't figure out what's the specific meaning of the representation obtained from neural networks, which is unaccountable or challenging to comprehend and will lead to an untrusty and irresponsible result.

%我们就想找一个又快又好解释，从以前的deep learning证明了学文本的低维表示是靠谱的，所以metric learn刚好就是这样，引入metric learning，且（以往是怎么用metric learning）

To tackle these, we aim to find a fast and interpretable approach for sentence matching.
There are several studies focused on learning low-dimensional representations of the data~\cite{salakhutdinov2009semantic}, which called metric learning~\cite{mikolov2013distributed} and even some of them combine it with some similarity metrics for ranking tasks \cite{kim2019deep}. 
Moreover, some researchers apply metric learning principles to design the loss function in information retrieval~\cite{bonadiman2019large} and question-answering tasks~\cite{bonadiman2017multitask}. But for the deep metric learning that they utilized, the neural network part still demands a lot of time. It hardly runs on a memory-limited device, together with high energy consumption. 
%我们工作是在text matching上提一个快速的方法。。。所以apply。。。

It is considering the unexplainable implications brought from neural networks, such as fairness or transparency, and the challenge of time-consuming. In this paper, we apply metric learning approaches to address the problems mentioned above. Because metric learning has an advantage in time and memory usage on large-scale and high-dimensional datasets compared with methods above. Here, metric learning finds a representation of the data that preserves these constraints that are placed by human-provided labels. Building on its success in learning ``label constraint preserving'' representations, or \emph{low-distortion embeddings}, we explore two \textsf{F}ast, \textsf{I}nterpretable, and \textsf{L}ow-rank \textsf{M}etric learning approaches, what we called \textsf{FILM}.

%看到效果metric learning能实现类似的结果，且快速且可解释性：线性的

Notably, we explore \textsf{FILM} methods on text matching tasks, which is also known as the semantic equivalence problem in the IR community~\citep{bogdanova2015detecting}. To be more specific, one based on an interpretable low-rank manifold optimization method. To solve this optimization problem,  we apply the Cayley transformation method with the Barzilai-Borwein step size. After being trained for this task, both are added to the kNN index for prediction for efficient retrieval. The input question is encoded and used as a query to the index, returning the top k most similar questions. We test our approaches on data from the Quora Challenge and SemEval-2017 Semantic Textual Similarity (STS) Task, which provide pairwise sentence similarity labels. 

%\footnote{\texttt{https://www.kaggle.com/c/quora-question-pairs}} 

%Our motivation is to investigate whether \textsf{FILM} approaches can perform as well as, if not better than, some ``black box'' approaches that are so popular these days. 
\begin{comment}

Our contributions are as follows:
\begin{itemize}
\item We design a deep metric learning method with a triplet loss function inspired by metric learning inspired by \cite{bonadiman2019large}; 
\item Inspired by a recently fast low-rank metric learning method~\cite{liu2019fast}, we proposed \textsf{FILM}---that relies on an interpretable linear model. Due to space constraints, we will focus on our main approach.
\end{itemize}

The rest of this paper is organized as follows. In Section \ref{secPrelim}, we provide a quick overview of metric learning. In Section \ref{secFILM} we present the interpretable \textsf{FILM} method. In Section \ref{secApp}, we summarize the Quora dataset and task, explain how \textsf{FILM} is applied to the task, and summarize our deep neural network approach. In Section \ref{secResults} we report some results. 
\end{comment}

\section{Preliminaries}
\label{secPrelim}
We begin with some definition and notation of metric learning, then we would follow the comparison to basic existing metric learning.

\subsection{Metric learning}
Metric learning is concerned with learning a similarity or distance function tuned to a particular task. In this paper, we introduce it as a similarity learning task. %but as we describe in this paragraph, the words``similarity'' and ``distance'' can be substituted interchangeably and depends strongly on the problem context. 
Given a pair of objects $\mathbf{x}$ and $\mathbf{x}'$, along with supervised information regarding an ideal similarity (resp., distance), the goal is to learn a function $\hat{s}(\mathbf{x},\mathbf{x}')$ (resp., $\hat{d}(\mathbf{x},\mathbf{x}')$), such that $\hat{s}$ (resp., $\hat{d}$) approximates the ideal similarity (resp., distance). 
Implicitly, $\hat{s}$ (resp., $\hat{d}$) learns a representation of $\mathbf{x}$---usually high-dimensional---under which the learned representation $\mathbf{y}=\hat{f}(\mathbf{x})$ contains exactly those features based on which the ideal similarity or distance is discriminating. Owing to this connection between metric learning and representation learning, metric learning is frequently used in regression and classification. 

In recent years, the rise of high-dimensional samples ($\mathbf{x}\in \mathbb{R}^D$ for $D$ large) and large sample sizes, along with various types of supervised information provided in real-world settings, have together spurred the development of scalable and flexible metric learning approaches. One type of metric learning~\cite{liu2015low,schultz2004learning,kwok2003learning,mason2017learning} utilizes the low-rankness encouraging norms (such as nuclear norm) as regularization, which relaxes the non-convex low-rank constrained problems to convex problems.  Another type of metric learning~\cite{mu2016fixed,harandi2017joint,cheng2013riemannian,huang2015projection,zhang2017efficient} considers the low-rank constrained space as Riemannian manifold. This type of method can obtain high-quality solutions for the non-convex low-rank constrained problems. However, the performance of these methods is still suffering from large-scale and high-dimensional datasets.

%\subsection{Metric Learning}
%\subsection{Low-Rank Metric Learning}
\subsection{Text Matching}
Our approaches are based on metric learning for the text matching task because of its flexibility toward the type of labels provided: given a collection of question samples $\{s_i\}$, whether the labels are (i) categories for each question $s_i$ (supervised), (ii) information about whether a pair of questions $\{s,s'\}$ are similar (semi-supervised), or (iii) information about whether for a triplet $\{s_i,s_j,s_k\}$, $s_i$ and $s_j$ are more similar than $s_i$ is to $s_k$ (semi-supervised), metric learning seeks to find a representation of the data that preserves these constraints that are placed by human-provided labels.  

\section{Our Approach}
\label{secFILM}
We propose a fast and interpretable metric learning approach for text matching, called \textsf{FILM}. \textsf{FILM} leverages the low rank of the data matrix to \emph{efficiently} find a projection of the high-dimensional data that still preserves high discriminative power, inspired by \cite{liu2019fast}. Given a training set with triplet labels, \textsf{FILM} learns a low-dimensional representation of the high-dimensional samples, such that in the low-dimensional space, samples originally more similar to one another are closer to one another in terms of some user-specified metric or similarity measure. We treat sentence matching as a manifold optimization problem. We carefully show how this problem formulation arises from a natural triplet loss minimization objective (Subsection \ref{subsec31}). In Subsection \ref{subsec32} we describe the algorithm for solving FILM.  

%\subsection{motivation}
\subsection{Formalization}
\label{subsec31}

Let the labeled dataset be $(\mathbf{X},\mathcal{T})$, where
 
\begin{itemize}
\setlength\abovedisplayskip{0cm}
\setlength\belowdisplayskip{0cm}
\item $\mathbf{X}=\begin{pmatrix}
\vline & \cdots & \vline \\
\mathbf{x}_1 &\cdots & \mathbf{x}_n \\
\vline & \cdots& \vline 
\end{pmatrix}\in \mathbb{R}^{D\times n}$ (feature-by-sample matrix)
\item $\mathcal{T}=\{(i,j,k): i~\text{more similar to}~j~\text{than}~k\}$ (\textbf{triplet constraint} labels)
\end{itemize}
 
Assume that $\mathbf{X}$ has low rank: $\text{rank}(\mathbf{X}) = r \leqslant \min(n,D)$. The main goal of FILM, like many metric learning approaches, is to learn a transformation $\mathbf{L}\in \mathbb{R}^{d\times D}$ such that under the representation $\mathbf{y}_i=\mathbf{L}\mathbf{x}_i$, $\text{sim}(\mathbf{y}_i,\mathbf{y}_j) > \text{sim}(\mathbf{y}_i,\mathbf{y}_k)$, where $\text{sim}(\cdot,\cdot)$ is a similarity measure chosen by the user. Additionally, \textsf{FILM} accomplishes this learning quickly by leveraging the low rank ($=r$) of $\mathbf{X}$. 

For the rest of the paper, we set
 \vspace{-0.06cm}
\begin{equation}
\setlength\abovedisplayskip{0cm}
\setlength\belowdisplayskip{0cm}
\text{sim}(\mathbf{y}_i,\mathbf{y}_j) = \mathbf{y}_i^T\mathbf{y}_j, \label{eq:31}
\end{equation}
 
which coincides with the cosine similarity of $\mathbf{y}_i$ and $\mathbf{y}_j$ when the two vectors have unit $\ell_2$ norm.

To find $\mathbf{L}$ such that Eq.~(\ref{eq:31}) holds most of the time for $(i,j,k)\in\mathcal{T}$, a natural way is to add a margin $m$, 
 
\begin{equation}
\setlength\abovedisplayskip{0cm}
\setlength\belowdisplayskip{0cm}
\text{sim}(\mathbf{y}_i,\mathbf{y}_j) \geqslant \text{sim}(\mathbf{y}_i,\mathbf{y}_k) +m,
\end{equation}
 
and to subsequently minimize the average hinge loss across all triplets:
 \vspace{-0.3cm}
\begin{equation}
\begin{aligned}
\setlength\abovedisplayskip{0cm}
\setlength\belowdisplayskip{0cm}
\min_{\mathbf{L}} \frac{1}{|\mathcal{T}|} \sum_{(i,j,k)\in\mathcal{T}} &\max(0, \text{sim}
(\mathbf{y}_i,\mathbf{y}_k)\\
&+ m - \text{sim}(\mathbf{y}_i,\mathbf{y}_j)).\label{eq:32}
\end{aligned}
\end{equation}
  \vspace{-0.1cm}
  
Each individual summand is a hinge loss, which is a convex approximation to the zero-one loss. This is the same loss function used by several deep neural network approaches to similarity prediction problems for image data and natural language data \cite{wang2014learning,bonadiman2019large}.  

Owing to the label set $\mathcal{T}$ and the sample size $n$ being large for typical datasets, evaluating Eq.~(\ref{eq:32}) is time-consuming without a GPU. Thus, \textsf{FILM} instead minimizes the average loss per \emph{subset} of triplets:
\begin{equation}
\setlength\abovedisplayskip{0cm}
\setlength\belowdisplayskip{0cm}
S=\sum_{(i,j,k)\in\mathcal{T}_i} (\text{sim}(\mathbf{y}_i,\mathbf{y}_k) + m - \text{sim}(\mathbf{y}_i,\mathbf{y}_j)),
\end{equation}

\vspace{-0.8cm}

\begin{equation}
\min_{\mathbf{L}} \sum_{i=1}^n \left[ \underbrace{\max\left(0, \frac{1}{|\mathcal{T}_i|+1}\times S \right)}_{\text{average hinge loss for sample}~i}\right] \label{eq:33}
\end{equation}
 
where $\mathcal{T}_i:=\{t=(t_1,t_2,t_3)\in \mathcal{T}: t_1=i\}$ is the set of triplets with $i$ as first component. (Note ``$|\mathcal{T}_i|+1$'' is used in the denominator instead of ``$|\mathcal{T}_i|$'' to avoid division by zero.) Following this, we will now reformulate the \emph{original} \textsf{FILM} problem---stated above as Problem (\ref{eq:33})---as a smooth optimization problem with the variable being a low-dimensional matrix lying on the Stiefel manifold.

\noindent\textbf{Step 1 (Restrict $\mathbf{L}$).} Let $\underset{D\times n}{\mathbf{X}}= [\underset{D\times r}{\mathbf{U}}, \underset{ r\times  r}{\mathbf{\Sigma}}, \underset{n\times  r}{\mathbf{V}}]$ be the SVD. Assume that $\mathbf{L}$ is the \emph{minimum norm least squares solution} of  $\mathbf{Y}=\mathbf{L}\mathbf{X}$, i.e., assume
\begin{equation}
\mathbf{L} =  \underset{d\times n}{\mathbf{Y}}~\underset{n\times r}{\mathbf{V}}~\underset{r\times r}{\mathbf{\Sigma}^{-1}}~\underset{r\times D}{\mathbf{U}^T}. \label{eq:34}
\end{equation}  
Through Eq.~(\ref{eq:34}), $\mathbf{L}$ is now defined in terms of $\mathbf{Y}$, and so Problem (\ref{eq:33}) can be reformulated as a problem involving the learned representations, $\mathbf{Y}=(\mathbf{y}_1,\ldots,\mathbf{y}_n)$, as the variable:

\vspace{-0.4cm}
\begin{equation}
\begin{aligned}
\hspace{-0.3cm}\min_{\mathbf{Y}} \sum_{i=1}^n [ \max (  
&0, \frac{1}{|\mathcal{T}_i|+1}\sum_{(i,j,k)\in\mathcal{T}_i} (\text{sim}(\mathbf{y}_i,\mathbf{y}_k)\\
&  + m - \text{sim}(\mathbf{y}_i,\mathbf{y}_j)))].  \label{eq:35}
\end{aligned}
\end{equation}

\noindent\textbf{Step 2 (Rewrite everything in terms of matrices).} To obtain a form of Problem (\ref{eq:35}) more amenable to mathematical optimization techniques, we shall define some matrices. For each $t=(t_1,t_2,t_3) \in\mathcal{T}$, let $\mathbf{C}^{(t)}$ be of size $n\times n$, with all entries $0$ except for $c^{(t)}_{t_2,t_1}=1$ and $c^{(t)}_{t_3,t_1}=-1$. Further, let $\mathbf{C}= \sum_{t\in\mathcal{T}}\mathbf{C}^{(t)}$. Observe that $\mathbf{Y}\mathbf{C}\in\mathbb{R}^{d\times n}$. Moreover, the $i$th column of $-\mathbf{YC}$, denoted by $\mathbf{\tilde{y}}_i$, is simply

\vspace{-0.1cm}
\begin{equation}
\setlength\abovedisplayskip{0cm}
\setlength\belowdisplayskip{0cm}
\mathbf{\tilde{y}}_i = \sum_{t\in\mathcal{T}_i} (-\mathbf{y}_j+\mathbf{y}_k).
\end{equation}
This is the sum of dissimilar samples minus the sum of similar samples for the triplet set $\mathcal{T}_i$. It helps to notice that 
\begin{small}
\begin{equation}
\begin{aligned}
\setlength\abovedisplayskip{0cm}
\setlength\belowdisplayskip{0cm}
\text{tr}\left(-\mathbf{Y}^T\mathbf{YC}\right) &= \sum_{i=1}^n \mathbf{y}_i^T\mathbf{\tilde{y}}_i \\
&= \sum_{i=1}^n \sum_{t\in\mathcal{T}_i} (-\mathbf{y}_i^T\mathbf{y}_j+\mathbf{y}_i^T\mathbf{y}_k)\\
&= \sum_{i=1}^n \sum_{t\in\mathcal{T}_i} (-\text{sim}(\mathbf{y}_i,\mathbf{y}_j)+\text{sim}(\mathbf{y}_i,\mathbf{y}_k)),
\end{aligned}
\end{equation}
 \end{small}
where the last equality follows from Eq.~(\ref{eq:31}). To finish the reformulation, further define 
\begin{itemize}
\setlength\abovedisplayskip{0cm}
\setlength\belowdisplayskip{0cm}
\item $\underset{n\times n}{\mathbf{T}}=\diag\left(\frac{1}{|\mathcal{T}_i|+1}\right)$,
\item $z_i = \frac{1}{|\mathcal{T}_i|+1} \mathbf{y}_i^T\mathbf{\tilde{y}}_i$ (the $i$th diagonal element of $-\mathbf{Y}^T\mathbf{YCT}$),
\item $\underset{n\times n}{\mathbf{\Lambda}}=\diag(\lambda(z_i+m))$, where $\lambda(x)=\begin{cases}
1 &\text{if}~x>0\\
0 & \text{otherwise}
\end{cases}$.
\end{itemize}

The objective function in Problem (\ref{eq:35}) is just $\sum_{i=1}^n(z_i+m)\lambda(z_i+m)=\sum_{i=1}^nz_i\lambda(z_i+m) + m\sum_{i=1}^n\lambda(z_i+m)$. With help from the previous notice, Problem (\ref{eq:35}) is equivalent to

\begin{equation}
\setlength\abovedisplayskip{0cm}
\setlength\belowdisplayskip{0cm}
\min_\mathbf{Y}\left\{ -\text{tr}\left(\mathbf{Y}^T\mathbf{Y}{\purple\mathbf{C}}{\purple \mathbf{T}}\mathbf{\Lambda}\right) + M(\mathbf{\Lambda})\right\}, \label{eq:36}
\end{equation}
where $M(\mathbf{\Lambda}) = m\cdot\text{tr}(\mathbf{\Lambda})$. The {\purple purple} objects in Problem (\ref{eq:36}) are constants depending on $\mathcal{T}$.

\noindent\textbf{Step 3 (Restrict $\mathbf{Y}$).} To leverage the low rank of $\mathbf{X}$, assume the factorization $\mathbf{Y}=\underset{d\times r}{\mathbf{B}}~\underset{r\times n}{\mathbf{V}^T}$. Plugging this factorization into the objective function of Problem (\ref{eq:36}) and applying the trace trick, we arrive at the following reformulation of Problem (\ref{eq:36}):

%\footnote{The authors of \textsf{FILM} frame this as a Theorem about such a factorization including all the possible minimum norm least squares solutions of $\mathbf{L}$, but this is confusing. A clearer statement is that, \emph{given that} $\mathbf{Y}$ admits such a factorization, then solving for $\mathbf{B}$ allows us to recover $\mathbf{L}$---assumed to be of the form of eq.~(\ref{eq:31})---since $\mathbf{L}=\mathbf{YV\Sigma}^{-1}\mathbf{U}^T=\mathbf{B\Sigma}^{-1}\mathbf{U}^T$.} 
 
\vspace{-0.2cm}
\begin{equation}
\setlength\abovedisplayskip{0cm}
\setlength\belowdisplayskip{0cm}
\min_\mathbf{\mathbf{B}}\left\{ -\text{tr}\left(\mathbf{B}^T\mathbf{B}{\purple \mathbf{V}^T}{\purple \mathbf{C}}{\purple \mathbf{T}}\mathbf{\Lambda}{\purple \mathbf{V}}\right) + M(\mathbf{\Lambda})\right\}. \label{eq:37}
\end{equation} 
 
Note: since $\mathbf{\Lambda}$ is a function of $\mathbf{Y}$ and $\mathbf{Y}$ is a function of $\mathbf{B}$, $\mathbf{\Lambda}$ is now viewed as a function of $\mathbf{B}$. Also, $\mathbf{V}$ depends on $\mathbf{X}$, so it is constant and coloured {\purple purple}.

In Problem (\ref{eq:37}), the objective function depends on $\mathbf{B}$ only through $\mathbf{B}^T\mathbf{B}$, because $\mathbf{\Lambda}$ depends on the $z_i$'s and each $z_i$ is a function of $\mathbf{Y}^T\mathbf{YCT}={\purple \mathbf{V}}\mathbf{B}^T\mathbf{B}{\purple \mathbf{V}^T}{\purple \mathbf{C}}{\purple \mathbf{T}}$. Assume that $r\geqslant d$ ($\text{rank}(\mathbf{X})$ is not smaller than the dimension of the intended representation). Since $\mathbf{B}^T\mathbf{B}$ is positive semi-definite with rank $d$, take the Jordan normal form $\mathbf{B}^T\mathbf{B}=\mathbf{PSP}^T$, where $\mathbf{S}\in \{\text{diag}(\mathbf{s}):\mathbf{s}\in\mathbb{R}^d_{+}\}$ is its set of eigenvalues and $\mathbf{P}\in \text{St}(d,r)$. Here,  

 \vspace{-0.2cm}
\begin{equation}
\setlength\abovedisplayskip{0cm}
\setlength\belowdisplayskip{0cm}
\text{St}(d,r)=\{\mathbf{P}\in\mathbb{R}^{r\times d}:\mathbf{P}^T\mathbf{P}=I_d\}
\end{equation}
 
is a \emph{Stiefel manifold}. Thus, Problem (\ref{eq:37}) is equivalent to minimizing a function of the pair $(\mathbf{P},\mathbf{s})$, with $\mathbf{P}$ lying on $\text{St}(d,r)$. 

\noindent\textbf{Step 4 (Regularize $\mathbf{s}$).} Let $\mathbf{K} = -\mathbf{V}^T\mathbf{CT\Lambda V}$. The objective function in Problem (\ref{eq:37}), written as a function of $(\mathbf{P},\mathbf{s})$, is given by 
  \vspace{-0.2cm}

\begin{equation}
\setlength\abovedisplayskip{0cm}
\setlength\belowdisplayskip{0cm}
f(\mathbf{P},\mathbf{s}) = \text{tr}(\mathbf{PSP}^T\mathbf{K}) + M(\mathbf{\Lambda}),
\end{equation}
where $\mathbf{\Lambda}$ now depends on $(\mathbf{P},\mathbf{s})$. We shall reduce the number of variables from two to one (only $\mathbf{P}$) by adding an $\ell_2$ penalty to $f$:
\vspace{-0.4cm}

\begin{equation}
\begin{aligned}
f_0(\mathbf{P},\mathbf{s}) & =  \text{tr}(\mathbf{PSP}^T\mathbf{K}) + \frac{1}{2}||\mathbf{s}||_2^2+M(\mathbf{\Lambda}) \\
					& =  \sum_{i=1}^d\left(\frac{1}{2}s_i^2 + s_i\mathbf{p}_i^T\mathbf{K}\mathbf{p}_i\right) + M(\mathbf{\Lambda}).
\end{aligned}
 \end{equation}

Treating $\mathbf{\Lambda}$ as constant, we see that the last expression is a sum of $d$ separate quadratic expressions involving each $s_i$. For any given $\mathbf{P}$, these quadratic expressions have a unique minimum, given by the following choices of $s_i$'s (which are assumed non-negative, since $\mathbf{B}^T\mathbf{B}$ is PSD):
 \vspace{-0.2cm}

\begin{equation}
\setlength\abovedisplayskip{0cm}
\setlength\belowdisplayskip{0cm}
s_i^\star  = \max\left(0,-\mathbf{p}_i^T\mathbf{K}\mathbf{p}_i\right).
\end{equation}

Plugging $\mathbf{s}^\star = (s_1^\star,\ldots,s_d^\star)$ back into $f_0$, we obtain 
 
\vspace{-0.7cm}

\begin{equation}
\begin{aligned}
\setlength\abovedisplayskip{0cm}
\setlength\belowdisplayskip{0cm}
f_1(\mathbf{P}) =& \frac{1}{2}\sum_{i=1}^d \left[\mathbf{p}_i^T\mathbf{K}\mathbf{p}_i \cdot \max\left(0,-\mathbf{p}_i^T\mathbf{K}\mathbf{p}_i\right)\right] \\
&+ M(\mathbf{\Lambda}),\label{eq:38}
\end{aligned}
\end{equation}

which depends only on $\mathbf{P}$. Therefore, Problem (\ref{eq:37}) is reformulated as the following problem. 
\begin{equation}
\setlength\abovedisplayskip{0cm}
\setlength\belowdisplayskip{0cm}
\begin{aligned}
\min\quad & f_1(\mathbf{P}) \\
\text{s.t.}\quad & \mathbf{P}=(\mathbf{p}_1,\ldots,\mathbf{p}_d)\in \text{St}(d,r)
\end{aligned} \label{eq:39}
\end{equation}
where $f_1(\cdot)$ is defined by Eq.~(\ref{eq:38}).

Note: the $\ell_2$ penalty, on top of allowing the derivation of a closed-form expression of $\mathbf{s}$ in terms of $\mathbf{P}$, also prevents the singular values of $\mathbf{B}$ from being too large. Moreover, $\mathbf{\Lambda}$ now depends only on $\mathbf{P}$---we solve for $\mathbf{P}$, then solve for $\mathbf{s}$ in terms of $\mathbf{P}$, and finally obtain $\mathbf{\Lambda}$ as a function of $\mathbf{PSP}^T$. 

\noindent\textbf{Step 5 (Smoothen the objective).} The objective function in Problem (\ref{eq:39}) does not have a continuous gradient. To remedy this, we replace $\max(0,x)$ by a smoothened surrogate, the negative log sigmoid function:
\begin{equation}
\setlength\abovedisplayskip{0cm}
\setlength\belowdisplayskip{0cm}
\mu(x) = -\log(\sigma(-x)),
\end{equation} 
where $\sigma(x) = 1/(1+ \exp(-x))$ is the sigmoid function and moreover $d\mu/dx= \sigma(x)$. Letting $k_i = -\mathbf{p}_i^T\mathbf{K}\mathbf{p}_i$, we substitute $\mu(x)$ for $\max(0,x)$ in the objective function of Problem (\ref{eq:39}), and reformulate Problem (\ref{eq:39}) as the following.
\begin{equation}
\setlength\abovedisplayskip{0cm}
\setlength\belowdisplayskip{0cm}
\begin{aligned}
\min\quad & f_2(\mathbf{P})=-\frac{1}{2}\sum_{i=1}^d k_i\cdot \mu(k_i) + M(\mathbf{\Lambda}) \\
\text{s.t.}\quad & \mathbf{P}=(\mathbf{p}_1,\ldots,\mathbf{p}_d)\in \text{St}(d,r)
\end{aligned} \label{eq:310}
\end{equation} 
Problem (\ref{eq:310}) is the smooth optimization problem we will solve. In terms of complexity, updating $\mathbf{\Lambda}$ requires $O(nrd)$ steps, updating $\mathbf{K}$ requires $O(nr^2)$ steps, updating $\nabla f_2(\mathbf{P})$ requires $O(r^2d)$ steps, updating $\mathbf{P}$ requires $O(rd^2) + O(d^3)$ steps, and updating $\mathbf{S}$ requires $O(r^2d)$ steps. 
%Thus, the overall complexity \emph{per iteration} of the algorithm (Steps 6 to 9) is linear in $n$, quadratic in $r$ and cubic in $d$.

\subsection{\textsf{FILM} Algorithm}
\label{subsec32}

To solve Problem (\ref{eq:310}), we apply the Cayley transformation method with Barzilai-Borwein step size. The Cayley transformation is a retraction, meaning that it approximates moving along a geodesic of the manifold up to first order accuracy. We provide an intuition for the method, justify its convergence, and describe the entire algorithm including its complexity and the recovery of $\mathbf{L}$. 

As an exposition we consider optimization on the Stiefel manifold for general functions. Let $\varphi:\text{St}(d,r)\rightarrow \mathbb{R}$ be a real-valued smooth function. Wanting to minimize $\varphi$, we write the Lagrangian 
 
\begin{equation}
\setlength\abovedisplayskip{0cm}
\setlength\belowdisplayskip{0cm}
\mathcal{L}(\mathbf{P,\Theta}) = \varphi(\mathbf{P}) -\frac{1}{2}\mathbf{\Theta}(\mathbf{P}^T\mathbf{P} - I_d),
\end{equation}
 
where $\underset{d\times d}{\mathbf{\Theta}}$ is the symmetric matrix of Lagrange multipliers.\footnote{It is symmetric because $\mathbf{P}^T\mathbf{P}-I_d$ is symmetric.} Setting the partial derivative of $\mathcal{L}$, $D_\mathbf{P}(\mathcal{L})$, to zero, we obtain the first order optimality condition $\nabla \varphi(\mathbf{P}) - \mathbf{P\Theta}=0$. Multiplying by $\mathbf{P}^T$ on both sides of this equation, we yield $\mathbf{\Theta}=\mathbf{P}^T\nabla \varphi(\mathbf{P})$. By symmetry of $\mathbf{\Theta}$, we obtain $\mathbf{\Theta}=\nabla \varphi(\mathbf{P})^T\mathbf{P}$, and plugging this back to the first order optimality condition produces
 
\begin{equation}
\setlength\abovedisplayskip{0cm}
\setlength\belowdisplayskip{0cm}
D_\mathbf{P}(\mathcal{L}) = \nabla \varphi(\mathbf{P}) - \mathbf{P}\nabla \varphi(\mathbf{P})^T\mathbf{P} = 0.
\end{equation}
 
Now define $\underset{r\times r}{\mathbf{A}}=\varphi(\mathbf{P})\mathbf{P}^T - \mathbf{P}\varphi(\mathbf{P})^T$, and observe that $\mathbf{A}\mathbf{P}=D_\mathbf{P}(\mathcal{L})$ if $\mathbf{P}$ were a critical point satisfying both first order optimality conditions $D_\mathbf{P}(\mathcal{L})=0$ and $D_\mathbf{\Theta}(\mathcal{L})=0$. 

Since $\mathbf{AP}$ is the gradient of the Lagrangian, a natural idea is to consider the following update:
 
\begin{equation}
\setlength\abovedisplayskip{0cm}
\setlength\belowdisplayskip{0cm}
\mathbf{P}_{k+1}(\tau) = \mathbf{P}_k -\tau \mathbf{A}\mathbf{P}_k,  \label{eq:311}
\end{equation}
 
where $\tau$ is a step size to be chosen later. However, Eq.~(\ref{eq:311}) doesn't guarantee $\mathbf{P}_{k+1}(\tau)\in \text{St}(d,r)$, and so an additional projection back to $\text{St}(d,r)$ is required if we wish to preserve the constraint at each iterate. (The work of \cite{absil2012projection} provides such projection methods.) Instead, for our problem we solve for 
 
\begin{equation}
\setlength\abovedisplayskip{0cm}
\setlength\belowdisplayskip{0cm}
\mathbf{P}_{k+1}(\tau) = \mathbf{P}_k -\tau\mathbf{A}\left(\frac{\mathbf{P}_k + \mathbf{P}_{k+1}(\tau)}{2}\right),\label{eq:312}
\end{equation}
which yields a closed form solution 
\begin{equation}
\setlength\abovedisplayskip{0cm}
\setlength\belowdisplayskip{0cm}
\mathbf{P}_{k+1}(\tau) = \underbrace{\left(I_r + \frac{\tau}{2} \mathbf{A}\right)^{-1}\left(I_r - \frac{\tau}{2} \mathbf{A}\right)}_{\text{Cayley transform}}\mathbf{P}_k.
\end{equation} 
 
This solution satisfies $\mathbf{P}^T_{k+1}(\tau)\mathbf{P}_{k+1}(\tau) = I_d$ whenever $\mathbf{P}_k^T\mathbf{P}_k=I_d$. By applying the Sherman-Morrison-Woodbury formula, it can also be written in the following form:
 
\begin{equation}
\setlength\abovedisplayskip{0cm}
\setlength\belowdisplayskip{0cm}
\mathbf{P}_{k+1}(\tau) = \mathbf{P}_k -\tau \mathbf{F} \left(I_{2d} + \frac{\tau}{2}\mathbf{G}^T\mathbf{F}\right)^{-1} \mathbf{G}^T\mathbf{P}_k,\label{eq:313}
\end{equation} 
 
where $\underset{r\times 2d}{\mathbf{F}} = [\nabla \varphi(\mathbf{P}_k),\mathbf{P}_k]$ and $\underset{r\times 2d}{\mathbf{G}}=[\mathbf{P}_k,-\nabla \varphi(\mathbf{P}_k)]$. This form facilitates faster computing if $2d \leqslant r$, since computing the matrix inverse expression now involves a matrix of dimension $2d\times 2d$ rather than $r\times r$. The approach of Eq.~(\ref{eq:312}) has been used to develop numerical algorithms for $p$-harmonic flows \cite{goldfarb2009curvilinear} and for optimizing weights in neural networks \cite{nishimori2005learning}.

Given the updated equation, now it remains to choose the step size $\tau$. We apply the \emph{Barzilai-Borwein (BB) method} \cite{barzilai1988two}, which says that $\tau$ is chosen by minimizing $g(\tau)=||\Delta \mathbf{P}_k - \tau \Delta(\nabla \varphi)||$, i.e., to have $\tau$ ``look like'' $[\text{Hess}_\varphi(\mathbf{P}_k)]^{-1} \cdot \nabla \varphi(\mathbf{P}_k)$ at step $k$ of the iteration. The BB method is a ``looking back'' approach that accelerates gradient methods \emph{at nearly no extra cost}, unlike traditional ``looking back'' line search approaches (e.g., the Armijo and Wolfe conditions).

\textbf{Barzilai-Borwein (BB) Method}.
Given the updated equation, now it remains to choose the step size $\tau$. We apply the \emph{Barzilai-Borwein (BB) method}, which says that $\tau$ is chosen by minimizing $g(\tau)=||\Delta \mathbf{P}_k - \tau \Delta(\nabla \varphi)||$, i.e., to have $\tau$ ``look like'' $[\text{Hess}_\varphi(\mathbf{P}_k)]^{-1} \cdot \nabla \varphi(\mathbf{P}_k)$ at step $k$ of the iteration. The BB method is a ``looking back'' approach that accelerates gradient methods \emph{at nearly no extra cost}, unlike traditional ``looking back'' line search approaches (e.g., the Armijo and Wolfe conditions). 

When optimizing on the Stiefel manifold, the BB method produces
\begin{small}
\begin{equation}
\tau_{k,1} = \frac{\text{tr}\left((\mathbf{P}_k - \mathbf{P}_{k-1})^T(\mathbf{P}_k - \mathbf{P}_{k-1})\right)}{|\text{tr}\left((\mathbf{P}_k - \mathbf{P}_{k-1})^T(\mathcal{F}(\mathbf{P}_k) - \mathcal{F}(\mathbf{P}_{k-1}))\right)| }
\end{equation}
\end{small}
or 
\begin{small}
\begin{equation}
\tau_{k,2} = \frac{|\text{tr}\left((\mathbf{P}_k - \mathbf{P}_{k-1})^T(\mathcal{F}(\mathbf{P}_k) - \mathcal{F}(\mathbf{P}_{k-1}))\right)|}{\text{tr}\left((\mathcal{F}(\mathbf{P}_k) - \mathcal{F}(\mathbf{P}_{k-1}))^T(\mathcal{F}(\mathbf{P}_k) - \mathcal{F}(\mathbf{P}_{k-1}))\right)},
\end{equation}
\end{small}
where $\mathcal{F}(\mathbf{P}) = \nabla \varphi(\mathbf{P}) - \mathbf{P}\nabla \varphi(\mathbf{P})^T\mathbf{P}$ (note $\mathcal{F}(\mathbf{P})= D_\mathbf{P}(\mathcal{L})=0$ if $\mathbf{P}$ is a critical point). 

\noindent\textbf{Justification of convergence.} The measure of convergence is $\varepsilon$-optimality of the gradient: $||\nabla \varphi(\mathbf{P^\star})||< \varepsilon\Rightarrow \mathbf{P}^\star$ is an approximate minimizer. The BB method by itself does not guarantee convergence, because the gradient is not monotone decreasing in the iterate. To remedy this, a non-monotone line search is incorporated into the step size calculation . 

\textbf{Full algorithm.} To apply the Cayley transformation with BB step size to solve Problem (\ref{eq:310}), we first compute the gradient of $f_2$, $\nabla f_2$. This is done by an approximation: we fix $\mathbf{\Lambda}$ and differentiate everything else in the objective function of Problem (\ref{eq:310}) w.r.t.~$\mathbf{P}$. By the chain rule, we obtain
 
\begin{equation}
\nabla f_2(\mathbf{P}) = -(\mathbf{K} + \mathbf{K}^T)\mathbf{P}\text{diag}(\mathbf{q}),\label{eq:314}
\end{equation}
 
where $\mathbf{q}$ is a vector with elements $q_i = -\frac{1}{2}(\mu(k_i) + k_i\sigma(k_i))$. By substituting RHS of Eq.~(\ref{eq:314}) for $\nabla\varphi$ in the update of Eq.~(\ref{eq:313}) with $\tau$ chosen by the BB method described earlier in this Subsection, we obtain an algorithm for solving Eq.(\ref{eq:310}).
%\footnote{In theory, the fact that $\mathbf{\Lambda}$ is fixed when computing the gradient implies that our algorithm does not provide an exact solution to the problem. However, on image and text datasets this algorithm has learned mappings $\mathbf{L}$ and representations $\mathbf{y}$ that preserve the triplet constraints and achieve high classification accuracy with the kNN class assignment rule. We also note that $\mathbf{\Lambda}$ is sparse, bounded and diagonal, so it is not surprising if the approximate gradient is close to the true gradient.}

\begin{algorithm}[t]
  \caption{\textsf{FILM}}
  \begin{algorithmic}[1]
  \STATE{\textbf{Input:}} Data matrix $\underset{D\times n}{\mathbf{X}}$, supervision matrix $\underset{n\times n}{\mathbf{C}}$ (obtained from $\mathcal{T}$), low-rank constraint $d$ (decided by user)
  \STATE Compute SVD: $\mathbf{X} = [\underset{D\times r}{\mathbf{U}}, \underset{ r\times  r}{\mathbf{\Sigma}}, \underset{n\times  r}{\mathbf{V}}]$
  \STATE Compute constant matrix ${\purple \mathbf{V}^T\mathbf{CT}}$
  \STATE Randomly initialize $\mathbf{P}\in\text{St}(d,r)$ and $\mathbf{S}\in \{\text{diag}(\mathbf{s}):\mathbf{s}\in\mathbb{R}^d_{+}\}$
  \REPEAT
  \STATE Update $\mathbf{\Lambda}$ (see \textbf{Step 2}) and $\mathbf{K}$ (see \textbf{Step 4})
  \STATE Update $\nabla f_2$ by Eq.~(\ref{eq:314})
  \STATE Update $\mathbf{P}$ by Eq.~(\ref{eq:313})
  \STATE Set $\mathbf{S}=\text{diag}(\mathbf{s}^\star)$ (see \textbf{Step 4})
  \UNTIL convergence
  \STATE{\textbf{Output:}} $\mathbf{L}=\sqrt{\mathbf{S}}\mathbf{P}^T\mathbf{\Sigma}^{-1}\mathbf{U}^T$
  \end{algorithmic}
\end{algorithm} 

\subsection{\textsf{FILM} for Matching}
\label{secApp}

\noindent\textit{Decision rule.} \textsf{FILM}, as described in Section \ref{secFILM}, transforms samples $\mathbf{x}_i$---which encode sentences---into low-dimensional vectors $\mathbf{y}_i$. However, the Quora task is to determine whether a given pair of sentences $(s,s')$ in the unlabeled test corpus are similar or dissimilar. Owing to \textsf{FILM} accomplishing successful label prediction typically when label decisions are made using the $k$-nearest neighbors (kNN) rule, we propose an analogous kNN approach to sentence pairwise similarity prediction. We call it the \textbf{Pairwise kNN} rule. Given sentences $(s,s')$ in validation or test set, assign them to be similar when their respective collection of $k$ nearest neighbours, identified by cosine similarity in the \textsf{FILM}-generated representation, contains the other.

We explain how $k$ is chosen in Step 6 as belows.
\noindent\textit{Execution.} Given the decision rule, we apply \textsf{FILM} to the Quora dataset as follows.
\begin{enumerate}
\item Split the labeled corpus into a training set $(\mathscr{S}_\text{train},\mathcal{P}_\text{train})$ and a validation set $(\mathscr{S}_\text{val},\mathcal{P}_\text{val})$. The split ratio is $4:1$. Run TF-IDF on $\mathscr{S}_\text{train}$ to obtain high-dimensional embedding/vectorization $\mathbf{X}_\text{train}$.
\item Run a clustering triplet generating method to obtain triplet constraints $\mathcal{T}_\text{train}$ from $\mathcal{P}_\text{train}$, which is as same as the triplet generation of deep metric learning described later.
\item Run the mini-batch version of \textsf{FILM} (\textsc{M-\textsf{FILM}}) on $(\mathbf{X}_\text{train},\mathcal{T}_\text{train})$ (to save memory).
\item Apply the learned TF-IDF embedding/vectorizer to $\mathscr{S}_\text{val}$ to obtain $\mathbf{X}_\text{val}$.
\item Apply the learned $\mathbf{L}$ from Step 4 to $(\mathbf{X}_\text{val},\mathcal{P}_\text{val})$, and choose the value of $k$ in Pairwise kNN that optimizes prediction performance (measured by cross-entropy loss).
\item Generate predictions on $\mathscr{S}_\text{test}$ by (i) applying the TF-IDF vectorizer to each sentence $s$ in $\mathscr{S}_\text{test}$ to obtain the corresponding embeddings $\mathbf{x}$, (ii) applying the learned $\mathbf{L}$ to each embedded sentence, (iii) applying Pairwise kNN with $k$ picked in Step 6 to assign similar (``1'')/not similar (``0'') to each sentence pair $(s,s')\in \mathscr{S}_\text{test}$.   
\end{enumerate}

\subsection{Deep metric learning with BERT}

Apart from \textsf{FILM}, we design a deep neural network with loss function similar to the average hinge loss of \textsf{FILM}. The setup consists of three separate feed-forward networks, which take in anchor samples, positive samples and negative samples. Here, anchor sample refers to a sentence $s$, positive sample refers to a sentence $s_+$ similar to $s$, and negative sample refers to a sentence $s_-$ dissimilar to $s$. The generation of such triplets $\{s,s_+,s_-\}$ is the same as in Step 3 of the Execution of \textsf{FILM} in the previous Subsection. We apply BERT \cite{devlin2018bert} to embed the sentences into vectors $\mathbf{x}$ for passing into the network.

Following the model in \cite{hoffer2015deep}, our three feed-forward networks have identical convolutional architectures, consisting of three convolutional and $2\times 2$ max-pooling layers, followed by a fourth convolutional layer. A non-linear \textit{ReLU} activation function is applied between two consecutive layers. Optimization is performed with the average triplet hinge loss function,

\begin{equation}
\begin{aligned}
\min_{W} \frac{1}{|\mathcal{T}|}\sum_{(i,j,k)\in\mathcal{T}} [&\text{sim}(\phi_W(\mathbf{x}_i),\phi_W(\mathbf{x}_k)) + m\\
&-\text{sim}(\phi_W(\mathbf{x}_i),\phi_W(\mathbf{x}_j))]_+
\end{aligned}
\end{equation}

\vspace{-0cm}\noindent and with weights $W$ shared across the three networks. Note: this loss function is identical to the average hinge loss across all samples described earlier (eq.~(\ref{eq:32}) of Section \ref{secFILM}).        
\section{Results}
Here, we focus on a recent dataset published by the QA website Quora.com containing over 400K annotated question pairs containing binary paraphrase labels. We evaluate our models on the Quora question paraphrase dataset which contains over 400,000 question pairs with binary labels. 

\begin{comment}
 \begin{table}[t]
 \setlength\abovedisplayskip{0cm}
\setlength\belowdisplayskip{0cm}
\centering
\vspace{0mm}
\begin{tabular}[]{l r}
%\begin{tabular}{l r r r r r r}
\hline
\textbf{Parameter} & \textbf{Setting}\\
\hline 
attention probs dropout prob & 0.1\\
hidden act  & gelu \\
hidden dropout prob & 0.1 \\
hidden size & 768\\
initializer range& 0.02\\
  intermediate size& 3072\\
  max position embeddings& 512\\
  num attention heads& 12\\
  num hidden layers& 12\\
  type vocab size& 2\\
vocab size& 30522\\
\hline
\end{tabular}
\caption{Experimental parameters Setting.}
\end{table}
\end{comment}

In Quora dataset, our goal is to retrieve similar questions from the question pool, based on the assumption that answers to similar question could answer new questions. The Quora cQA Challenge can be summarized as follows.

\begin{itemize}
\item Given: Labelled corpus $(\mathscr{S},\mathcal{P})$, where $\mathscr{S}$ is a collection of sentence pairs $(s,s')$ and $\mathcal{P}$ is a collection of similarity tags $\text{sim}(s,s')\in\{0,1\}$. Note that $|\mathscr{S}| = 808580/2 = 404290$.
\item Task: Train a prediction model that takes in a sentence pair $(s_0,s_0')$  and returns a similarity tag $\hat{\text{sim}}(s_0,s_0')$. 
\item Evaluation: Performance on an unlabelled corpus provided by Quora, $\mathscr{S}_\text{test}$. Note that $|\mathscr{S}_\text{test}|=2345795$.
\end{itemize}

\vspace{-0.4cm}

\begin{table}[htbp]
\centering
\vspace{0mm}
\small
\begin{tabular}[]{l r}
%\begin{tabular}{l r r r r r r}
\hline
\textbf{Models} & \textbf{Accuracy(\%)}\\
\hline 
Siamese CNN & 79.60\\
Multi-Perspective-CNN  & 81.38 \\
Siamese-LSTM & 82.58 \\
Multi-Perspective-LSTM & 83.21\\
L.D.C. & 85.55\\
BiMPM-w/o-tricks & 85.88\\
BiMPM-Full & 88.17\\
FILM & 86.83\\
FILM-BERT & 88.09\\
\hline
\end{tabular}
\caption{Performance on Quora Question
Dataset.}
\label{table:Datasetstatistics} 
\end{table}

\begin{table}[htbp]
\centering
\vspace{0mm}
\small
\begin{tabular}[]{l r}
%\begin{tabular}{l r r r r r r}
\hline
\textbf{Models} & \textbf{Pearson Correlation}\\
\hline 
LexVec & 55.8 \\
FastText & 53.9\\
GloVe & 40.6\\
Word2vec  & 56.5 \\
FILM  & 57.68 \\
\hline
\end{tabular}
\caption{Performance on STS
Dataset.}
\label{table:Datasetstatistics2} 
\end{table}

We use the same data and split as \cite{wang2017bilateral}, with 10,000 question pairs each for development and test, who also provide preprocessed and tokenized question pairs.4 We duplicated the training set, which has approximately 36\% positive and 64\% negative pairs, by adding question pairs in reverse order (since our model is not symmetric).

Submissions are evaluated on the log loss between the predicted values and the ground truth. By applying our \textsf{FILM}, we obtain a log loss score of 0.32141 from a single classifier (trained less than 20 minutes). The loss of the model decreased rapidly in the initial epochs but flattened out into a plateau as the number of training steps increased which is shown in Fig. 2. 

\vspace{-0.2cm}
\begin{figure}[htbp]

\centering

\includegraphics[scale=0.3]{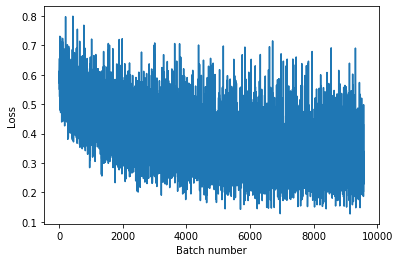}

\caption{Training loss is decreased when the numbers of training steps are increasing.}

\end{figure}
\begin{figure}[htbp]
\centering
\includegraphics[scale=0.3]{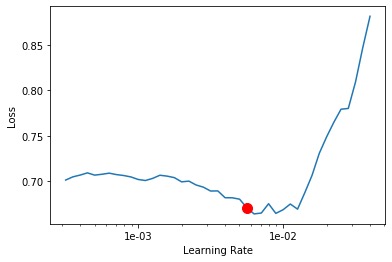}
\caption{Find a best min numerical gradient of 5.62E-03.}
\end{figure}

\begin{figure}[htbp]
\centering
\includegraphics[scale=0.3]{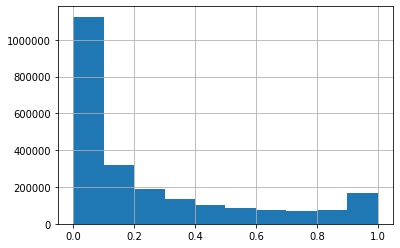}
\caption{Log loss of different sample in test dataset with hist() function.}
\end{figure}

\vspace{-0.2cm}

TF-IDF produced high-dimensional vectorizations $\mathbf{x}\in\mathbb{R}^{78113}$ of the training sentences. Setting $d=100$ and running \textsf{FILM} on the training set, we obtained a mapping $\mathbf{L}\in\mathbb{R}^{100\times 78113}$. We ran kNN for values of $k$ ranging from $1$ to $55$. For each value of $k$, we compute the cross-entropy loss (CE) and four other performance metrics: the True Positive Rate (TPR, defined as number of correctly classified similar pairs), True Negative Rate (TNR, defined as number of correctly classified dissimilar pairs), False Positive Rate (FPR) and False Negative Rate (FNR). We find that $k=1$ achieves the lowest CE, and the FNR does not decrease significantly as $k$ increases (see Figure2 below). Even though $k=1$ produces the lowest CE, we pick $k=36$ to run \textsf{FILM} on the test set, since there is an ``elbow'' in the CE for $k\geqslant 37$.       

\vspace{-0.4cm}
\section{Analysis}

After running the deep neural network, we find that our approaches have different strengths and weaknesses (Table \ref{tab:1}). \textsf{FILM} is executable on a CPU, but we had to run the black box model on a GPU.

\vspace{-0.5cm}

\begin{figure}[h]
\setlength\abovedisplayskip{0cm}
\setlength\belowdisplayskip{0cm}
\centering
\begin{minipage}[t]{0.4\textwidth}
\centering
\includegraphics[width=0.7\textwidth]{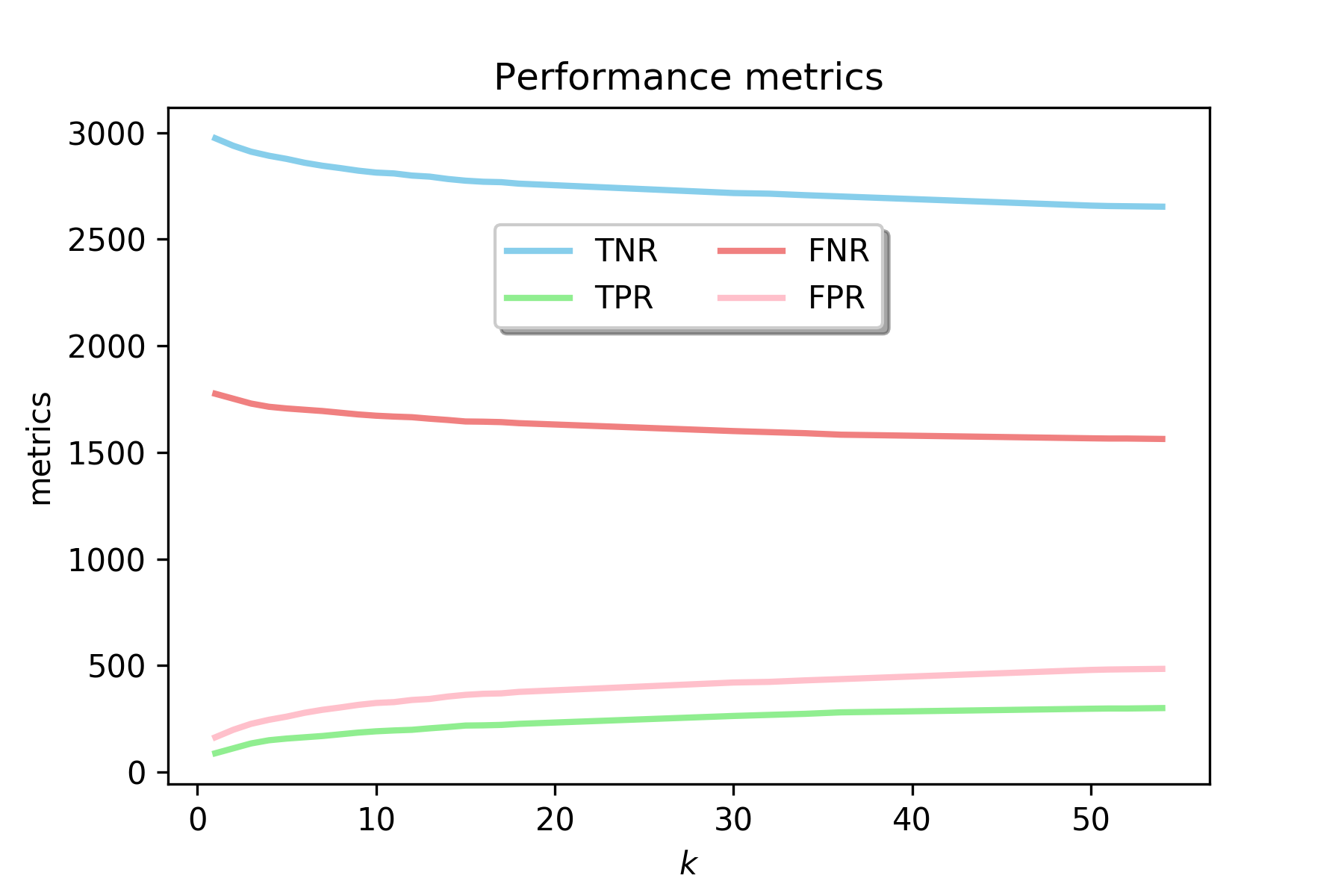}
\caption{Metrics for performance of \textsf{FILM} on the validation set, $(\mathscr{S}_\text{val},\mathcal{P}_\text{val})$, as described in FILM Algorithm Step 6 of Section 3.2. (Top) TPR, TNR, FPR, FNR. (Bottom) Cross-entropy loss.}
\end{minipage}

\begin{minipage}[t]{0.4\textwidth}
\centering
\includegraphics[width=0.7\textwidth]{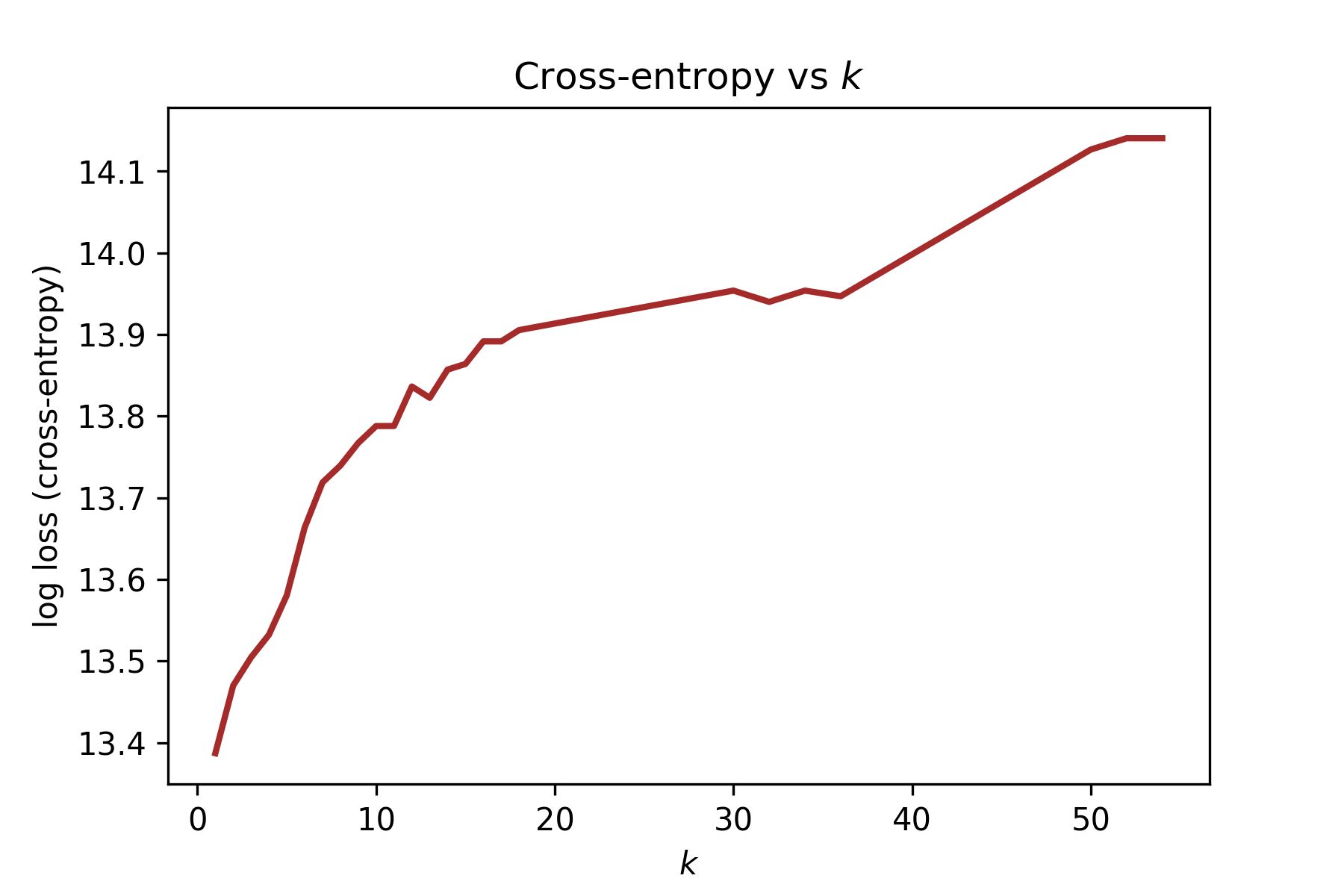}
\end{minipage}
\label{fig:1}
\end{figure}

\begin{figure}[h]
\centering
\includegraphics[scale=0.3]{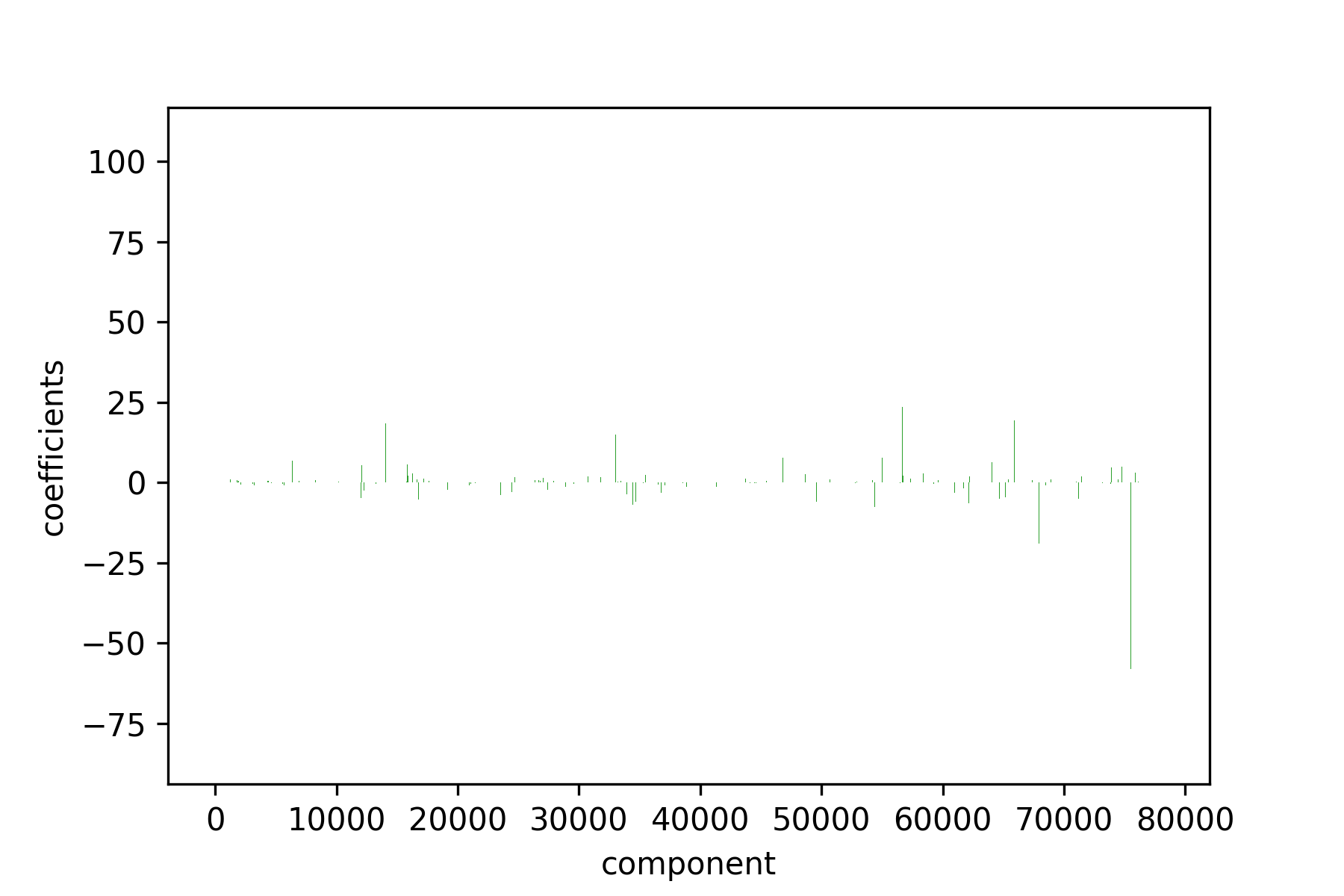}
\caption{The $78113$ components of the first row of $\mathbf{L}$, which correspond to coefficients of the linear combination of $\mathbf{x}$ that determines the first component of the representation $\mathbf{y}$: $\mathbf{l}_1\mathbf{x}=y_1$. }
\label{f}
\end{figure}

Finally, we find that the function learned on the black box model, $\phi$, lacks interpretability, whereas the mapping learned by \textsf{FILM}, $\mathbf{L}$, has a natural interpretation, as showed in Figure \ref{f}. Here, we observe that $\mathbf{l}_1$ has identified a small subset of features of $\mathbf{x}$. By comparing the nonzero indices of $\mathbf{l}_1$ with the corresponding word features in the TF-IDF embedding, we can identify if $\mathbf{l}_1$ has picked up any meaningful signal.   

The main unresolved problem of deep metric learning we found is overfitting. We can do detailed data analysis in the future to analyze the influence of selection bias and explore ways to select representative validation datasets. So the future work is to make metric learning more efficient to generalize model better. Also, it is hard to train, but we optimize it with N-Pair Loss. Moreover, the Quora dataset has a selection bias, which hurt the generalization performance of trained models and gives untrustworthy evaluation results. In other words, such train dataset leads to overfitting problem. In the future, we hope to propose frameworks that can improve the generalization ability of trained models, and give more trustworthy evaluation results for real-world adoptions.

\begin{table}[htb]
\label{tab:1}
\hspace{-0cm}\begin{tabular}{|c|c|c|}
\hline
\textbf{Characteristic}                         & \textbf{Black Box Model} & \textbf{\textsf{FILM}}         \\ \hline
\textit{Memory Usage}                           & Requires GPU             &  on CPU         \\ \hline
\textit{Speed}                                  & $80$ minutes       & $<30$ mins \\ \hline
\textit{Performance}                            & SOTA                  & Strong                      \\ \hline
\multicolumn{1}{|l|}{\textit{Interpretability}} & Weak                     & Strong                 \\ \hline
\end{tabular}
\caption{Summary of strengths and weaknesses of the neural network model and \textsf{FILM} based on application to text matching data.}
\end{table}

\vspace{-0.2cm}

\section{Conclusion}
\label{secConclusion}
We investigated text matching, a core task in information retrieval and semantic analysis. We introduced the notation and definition of metric learning, and how it can be applied to text matching. Then, we explored \textsf{FILM} (\textsf{F}ast, \textsf{I}nterpretable and \textsf{L}ow-rank \textsf{M}etric learning), which aim to reduces the time cost and memory usage, also save energy consumption. In order to solve this task efficiently, \textsf{FILM} combined with a fast approximate k nearest neighbour search index.
Compare to neural models, our method also has advantage in time and memory usage on large-scale and high-dimensional datasets.

\bibliography{aacl-ijcnlp2020}
\bibliographystyle{acl_natbib}

\end{document}